# Fast-UAP: An Algorithm for Speeding up Universal Adversarial Perturbation Generation with Orientation of Perturbation Vectors

**Jiazhu Dai**  
Shanghai University

**Le Shu**  
Shanghai University

## Abstract

Convolutional neural networks (CNN) have become one of the most popular machine learning tools and are being applied in various tasks, however, CNN models are vulnerable to universal perturbations, which are usually human-imperceptible but can cause natural images to be misclassified with high probability. One of the state-of-the-art algorithms to generate universal perturbations is known as UAP. UAP only aggregates the minimal perturbations in every iteration, which will lead to generated universal perturbation whose magnitude cannot rise up efficiently and cause a slow generation. In this paper, we proposed an optimized algorithm to improve the performance of crafting universal perturbations based on orientation of perturbation vectors. At each iteration, instead of choosing minimal perturbation vector with respect to each image, we aggregate the current instance of universal perturbation with the perturbation which has similar orientation to the former so that the magnitude of the aggregation will rise up as large as possible at every iteration. The experiment results show that we get universal perturbations in a shorter time and with a smaller number of training images. Furthermore, we observe in experiments that universal perturbations generated by our proposed algorithm have an average increment of fooling rate by 9% in white-box attacks and black-box attacks comparing with universal perturbations generated by UAP.

## I. INTRODUCTION

Deep learning has become the predominant machine learning method and showed outstanding performance in diverse tasks, e.g. self-driving cars [17], surveillance [18], malware detection [19], drones and robotics [20], [21], and voice command recognition [22]. Though deep learning has received much success, recent studies has revealed the vulnerability of deep learning.

Convolutional neural networks (CNN) are one category of the outstanding algorithms of deep learning and has performed state-of-the-art accuracy in image classification tasks. However, CNN models have been recently found vulnerable to well-designed input samples [1]-[8], [15]. These samples can easily fool a well-performed CNN model with little perturbations imperceptible to humans, which called adversarial perturbations. (e.g. Fig. 1). Szegedy *et al.* [1] first discovered the intriguing weakness of CNN models and introduced adversarial perturbations against them. The profound implications of their research have triggered a wide interest of researchers in adversarial attacks and defenses for CNN models, more methods to generate adversarial perturbations have been proposed [2]-[8], [15].

Though most adversarial perturbations are image specific (e.g. [1]-[7]), an image-agnostic adversarial perturbation called universal adversarial perturbation has been proposed by Moosavi-Dezfooli *et al.* [8], which is very small perturbation vector that causes a set of natural images to be misclassified with high probability. The algorithm proceeds iteratively over a set of training images and gradually aggregates perturbation vectors for each image to builds the universal perturbation. At each iteration, the minimal perturbation vector that sends the current perturbed image to the decision boundary of the classifier is computed and aggregated to the current instance of the universal perturbation. However, the minimal perturbations are not the best choice to craft universal perturbations, because orientation of current minimal perturbation vector and that of the current instance of the universal perturbation vector has impact on the magnitude of their new aggregation result. For example, if they have similar orientation, the magnitude of their aggregation will be larger, and the magnitude of their aggregation will be smallest if they have opposite orientation. As a result, the magnitude of constructed universal perturbations may not rise up efficiently in every iteration, it will slow down the process of generation. Motivated by this analysis, in this paper, we propose an optimized algorithm to craft universal perturbations faster. At each iteration, instead of choosing minimal perturbation vector, we aggregate the current instance of the universal perturbation with the perturbation which has similar orientation to the former so that the magnitude of the aggregation will rise up as large as possible at every iteration. As a result, we can get a universal perturbation whose magnitude rise up as soon as possible. Besides, we empirically observe that universal perturbations generated by our proposed algorithm have better fooling performance in both white-box attacks and black-box attacks with an average increment of fooling rate by 9%.

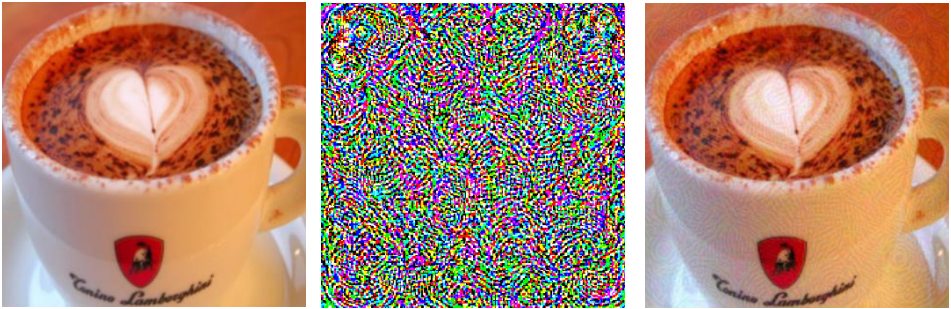

**FIGURE 1.** An example of adversarial perturbation. Left: an original natural image classified as espresso with the highest confidence by GoogLeNet model. Center: an adversarial perturbation. Right: an adversarial sample craft by adding the center adversarial perturbation to the left natural image and it classified as pillow with the highest confidence by GoogLeNet model.

The rest of the paper is organized as follows: we review the related work in section II. In section III, we demonstrate orientation of perturbation vector has impact on the generated universal perturbation and illustrate our inspiration, then, we propose our approach to speed up universal perturbation generation. In section IV, we present our experiment and show results in detail. Finally, we conclude our work and draw the future research direction in section V.

## II. RELATED WORK
Despite the impressive performance of CNN models on image classification tasks, these models were shown to be highly vulnerable to adversarial perturbations. These perturbations are divided into two categories: one is image-specific perturbations, which are sought to fool a specific image on a CNN model. The other is universal perturbations, which are image-agnostic and can fool a CNN model on a set of images. Szegedy et al. [1] proposed the first algorithm to generate image-specific perturbations, called L-BFGS, which generates perturbation by performing a single step computation. However, L-BFGS used an expensive linear search method to find the optimal value, it cost too much time to get a perturbation. Goodfellow et al. [2] proposed a fast method to generate perturbations, called Fast Gradient Sign Method (FGSM), which only performs once optimization along the sign of gradient of each pixel. Soon after, there were some improvements of FGSM, Rozsa el al. [9] proposed Fast Gradient Value method, Tramer et al. [10] proposed RAND-FGSM and Dong et al. [11] applied momentum to FGSM to generate perturbations iteratively, for iterative algorithm are harder to defend comparing with one step computation algorithm [12]. And there are other outstanding methods to generate image-specific perturbations [13], [14], [3]-[7]. Particularly, Liu et al. [7] proposed perturbations that can perform black-box attacks, which generated with a limited knowledge of the model, such as its architecture rather than the model parameters.

Moosavi-Dezfooli et al. [8] first designed an iterative algorithm to generate universal perturbations which cause target CNN model return incorrect predictions with high probability on a set of images while still remains quasi-imperceptible to the human eye. The algorithm is known as UAP which is based on DeepFool [3]. DeepFool generates minimal perturbations that gradually pushing every image to the decision boundary for their classes, and all the perturbations generated for each image are accumulated to a universal perturbation. Mopuri et al. [15] proposed data-free UAP, which does not craft universal perturbations with natural images directly. Instead, they first generate class impressions which are generic representations of the images belonging to every category, and then present a neural network based generative model to craft universal perturbations with class impressions using data-driven objectives. Compared with UAP, our proposed Fast-UAP speeds up universal perturbation generation based on orientations of perturbation vectors, which aggregates additive perturbations with similar orientation to current instance of aggregations in every iteration. Besides, it is observed in experiment that the universal perturbations generated by Fast-UAP achieve higher average fooling rates in both white-box attacks and black-box attacks.

## III. THE FAST-UAP ALGORITHM

### A. OVERVIEW

UAP adopts an iterative method to generate universal perturbations to fool CNN models on image classification tasks. Let $X = \{x_1, x_2, x_3, ..., x_n\}$ be a set of natural images, UAP seeks a universal perturbation $v$ that can fool most images in $X$ on target classifier $C(\cdot)$. For each image $x_i$, if current constructed universal perturbation $\bar{v}$ does not fool $C(\cdot)$ on $x_i$, a minimal perturbation $\Delta_{v_i}$ is computed and aggregated to $\bar{v}$, here $\Delta_{v_i}$ is the perturbation with the smallest magnitude to push image $x_i$ out of decision boundary of its estimated class. Which can be expressed as:

$$\bar{v} \leftarrow \bar{v} + \Delta_{v_i}.$$

UAP computes $\Delta_{v_i}$ with respect to $x_i$ based on DeepFool [3] by solving the following optimization problem:

$$\Delta_{v_i} \leftarrow \arg\min_{v_i} ||v_i||_2 \text{ s.t. } C(x_i + \bar{v} + v_i) \neq C(x_i).$$

Where $C(x_i)$ denotes the output of target classifier $C(\cdot)$ when performing $x_i$ as an input.

Fast-UAP takes similar process to generate universal perturbations. Instead of aggregating minimal perturbation in every iteration, Fast-UAP aggregates additive perturbations with a similar orientation to current instance of universal perturbations.

### B. FAST-UAP

#### 1) INSPIRATION

Given two vectors $\boldsymbol{a}$ and $\boldsymbol{b}$, the magnitude of $\boldsymbol{a}$ add $\boldsymbol{b}$ is computed as follows:

$$||\boldsymbol{a} + \boldsymbol{b}||_2 = \sqrt{||\boldsymbol{a}||_2^2 + ||\boldsymbol{b}||_2^2 + 2||\boldsymbol{a}||_2 ||\boldsymbol{b}||_2 \cos <\boldsymbol{a}, \boldsymbol{b}>}. \quad (1)$$

Where $\cos <\boldsymbol{a}, \boldsymbol{b}>$ denotes cosine of angle between $\boldsymbol{a}$ and $\boldsymbol{b}$. Thus, to get a larger magnitude of sum of $\boldsymbol{a}$ and $\boldsymbol{b}$, the angle between $\boldsymbol{a}$ and $\boldsymbol{b}$ should be as small as possible, which means both $\boldsymbol{a}$ and $\boldsymbol{b}$ have more similar orientation.

Inspired by this, we now demonstrate the slow universal perturbation generation in UAP. As described above, UAP proceeds by aggregating additive perturbations with the smallest magnitude to push successive images out of decision boundaries of their target classifier's estimated classes. However, it ignores the influence of perturbation vectors' orientations. Though these minimal perturbations can push corresponding images out of decision boundaries, the magnitude of their aggregation does not rise up efficiently in successive iterations, as depicted in Fig. 2a. There is a set of images $X = \{x_1, x_2, x_3\}$ with corresponding class label $\{\mathcal{R}_1, \mathcal{R}_2, \mathcal{R}_3\}$ to generate a universal perturbation. When UAP proceeds over images $x_1, x_2, x_3$ in sequence, the minimal perturbation $\Delta_{v_i}$ which sends the corresponding perturbed image $x_i + \Delta_{v_i}$ to the decision boundary is computed, and aggregated to the current instance of universal perturbation. However, the orientations of the minimal perturbations have impact on the aggregations' magnitude growth as shown in (1). When the minimal perturbations are added, magnitude of current universal perturbation may be smaller than that in past iteration because the minimal perturbations have arbitrary orientations which likely lead to a drop in universal perturbation's magnitude. For example, in Fig. 2a, when adding $\Delta_{v_2}$, the current perturbation $\bar{v}$ has a smaller magnitude than $\Delta_{v_1}$ in past iteration. Because of the drop, the magnitude of generated universal perturbation $v$ does not rise up to the level that is large enough to push the three images out of their decision boundaries. This will lead to a slow universal perturbation generation.

To ensure the magnitudes of generated universal perturbations rise up efficiently in every iteration, when proceeding over $x_1, x_2, x_3$ in sequence to generate a universal perturbation, extra perturbation $\Delta_{v_i}$ is also computed to send the corresponding perturbed image $x_i + \Delta_{v_i}$ to the decision boundary of target classifier, here we choose extra perturbation $\Delta_{v_i}$ which have similar orientation to current constructed universal perturbation rather than the minimal one, as depicted in Fig. 2b. The orientations of current universal perturbations in every iteration are reference orientations for choosing extra perturbations, and it is initialized with the first generated extra perturbation $\Delta_{v_1}$ with respect to $x_1$. If the extra perturbation has a more similar orientation to current universal perturbation, the magnitude of their aggregation will increase. For example, in Fig. 2b, the extra perturbation $\Delta_{v_2}$ has a similar orientation to current universal perturbation $\Delta_{v_1}$ and when it is added, the aggregation $\bar{v}$ has a large magnitude growth compared with $\Delta_{v_1}$ in past iteration. As a result, the magnitude of generated universal perturbation $v$ can be large enough to push all three images out of their decision boundaries. This will speed up universal perturbation's magnitude growth and lead to a fast universal perturbation generation.

#### 2) ALGORITHM

Given image $x$ in $\mathbb{R}^d$ with distribution $\mu$ and a target classifier $C(\cdot)$, we seek a quasi-imperceptible universal perturbation $v \in \mathbb{R}^d$ that satisfy:

$$C(x) \neq C(x + v) \text{ for most } x \text{ in } \mu.$$

Where $\mu$ refers to the distribution of natural images and $C(x)$ is the output of target classifier $C(\cdot)$ when performing $x$ as an input. Our training set is a set of images $X = \{x_1, x_2, x_3, ..., x_n\}$ and target classifier is a CNN model, the algorithm iteratively proceeds over images $x_i$ and gradually constructs the universal perturbation. In more detail, for every $x_i$, if the combination of $x_i$ and current constructed universal perturbation $\bar{v}$ does not make a label change on the target classifier, an extra perturbation $\Delta_{v_i}$ with respect to $x_i$ that has a similar orientation to $\bar{v}$ is crafted, or this iteration will be skipped. The extra perturbation is generated by solving the following optimization problem:

$$\Delta_{v_i} \leftarrow \arg\max_{v_i} \cos <v_i, \bar{v}> \text{ s.t. } C(x_i + \bar{v} + v_i) \neq C(x_i).$$

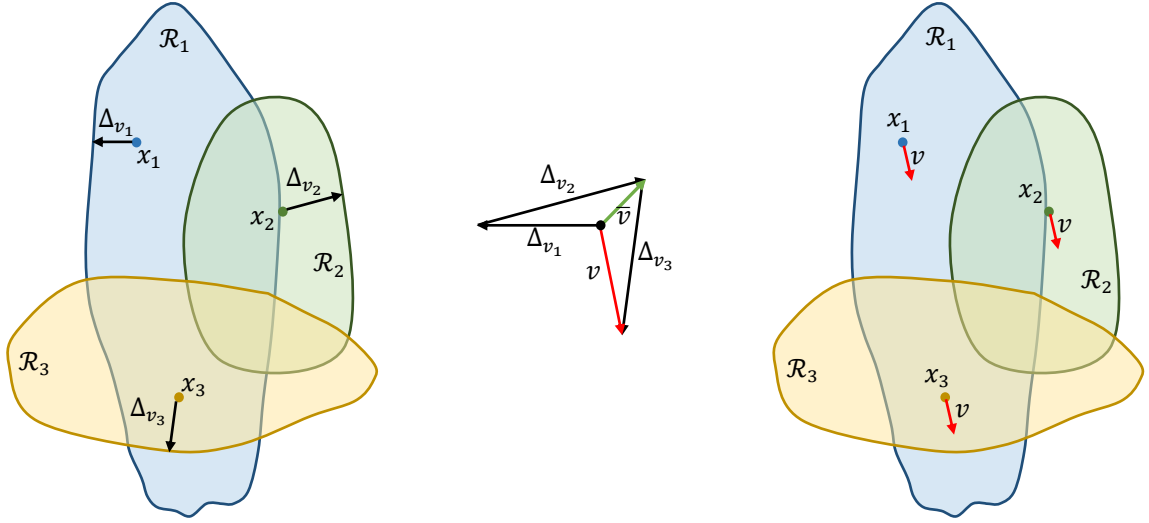

**(a)** Schematic representation of universal perturbation's slow magnitude growth in UAP. When adding perturbation $\Delta_{v_2}$, current universal perturbation $\bar{v}$ has a magnitude drop compared with $\Delta_{v_1}$ and the magnitude of generated universal perturbation $v$ does not get a sufficient growth to push the images $x_1$, $x_2$, and $x_3$ out of their decision boundaries.

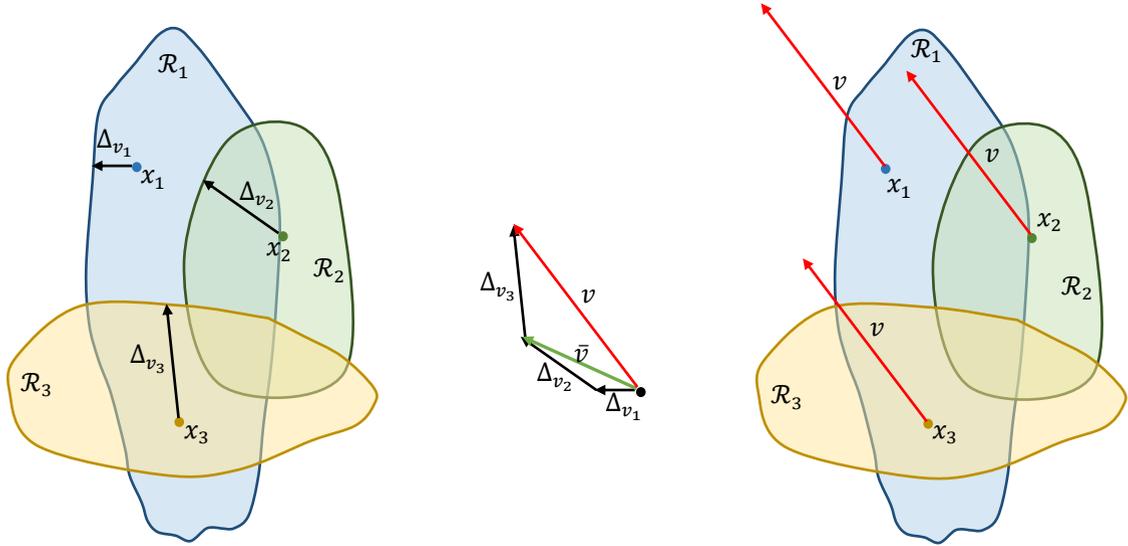

**(b)** Schematic representation of Fast-UAP to speed up universal perturbation's magnitude growth. Perturbation $\Delta_{v_2}$ has similar orientation to $\Delta_{v_1}$ and $\Delta_{v_3}$ has similar orientation to $\bar{v}$. As a result, the magnitudes of $\bar{v}$ (compared with $\Delta_{v_1}$) and $v$ (compared with $\bar{v}$) rise up efficiently and the generated universal perturbation $v$ can push the images $x_1$, $x_2$, and $x_3$ out of their decision boundaries because of its large magnitude.

**FIGURE 2.** Schematic representations of universal perturbation's slow magnitude growth in UAP and our improvement to overcome this influence. In these illustrations, points $x_1$, $x_2$, and $x_3$ refer to three images belonging to classification $\mathcal{R}_1$, $\mathcal{R}_2$, and $\mathcal{R}_3$ respectively and classification regions $\mathcal{R}_i$ are shown in different colors. Algorithm proceeds over $x_1, x_2, x_3$ in order. In both Fig. 2a and Fig. 2b: left: illustration of perturbation vectors $\Delta_{v_1}$, $\Delta_{v_2}$, and $\Delta_{v_3}$ to push the images $x_1$, $x_2$, and $x_3$ out of their decision boundaries respectively. Center: illustration of addition of the perturbation vectors $\Delta_{v_1}$, $\Delta_{v_2}$, and $\Delta_{v_3}$ (perturbation vectors in Fig. 2a are scaled up for a better visibility). Right: illustration of attack ability of generated universal perturbation on the images $x_1$, $x_2$, and $x_3$.

Where $\Delta_{v_i}$ is the extra perturbation that has similar orientation to current universal perturbation $\bar{v}$, and it can push image $x_i$ out of decision boundary of its estimated class but is not always the minimal one. Note that if two vectors have a bigger cosine value, they possess a more similar orientation. We choose a random image $x_0$ in $X$, and take the orientation of minimal perturbation computed with respect to $x_0$ as initialization of universal perturbation. At the end of each iteration, to make universal perturbation being quasi-imperceptible to the human eye, we limit the pixel values of universal perturbation within $[-10, 10]$, which is a $\ell_\infty$ norm restriction. So, the current constructed universal perturbation will be clipped once an extra perturbation is added. At the end of every pass on training set $X$, the algorithm will check

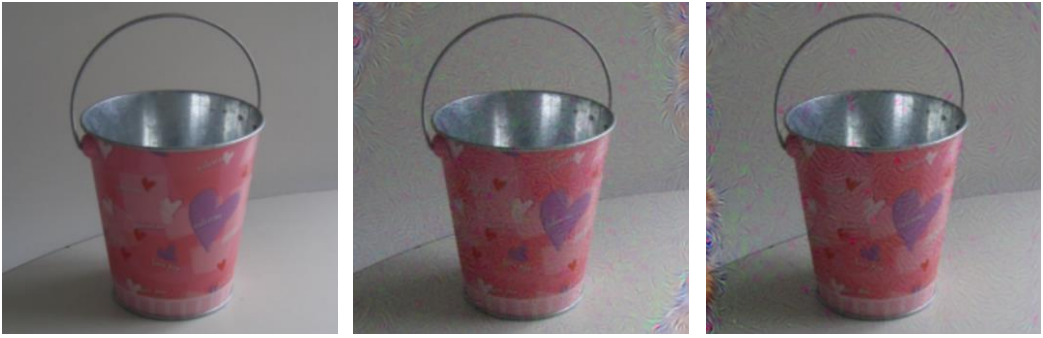

**FIGURE 3.** Chromatic aberration in perturbed images when adding $\ell_2$ norm to restrict universal perturbations generated by Fast-UAP. Left: an original image. Center: a perturbed image by adding universal perturbation generated for DenseNet-121 model. Right: a perturbed image by adding universal perturbation generated for ResNet-50 model. The chromatic aberrations are easily seen at the edge of the center and the right images.

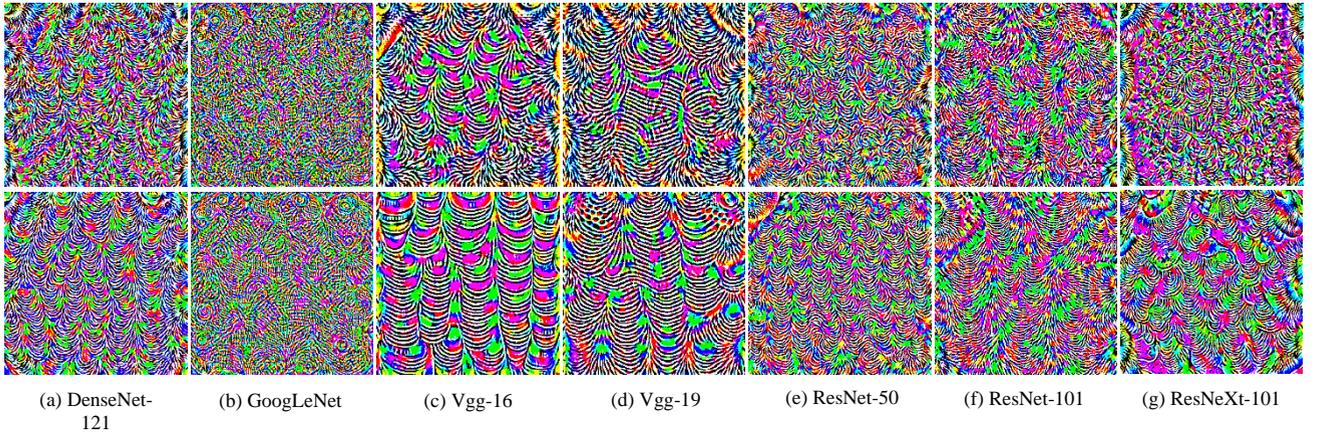

(a) DenseNet-121  (b) GoogLeNet  (c) Vgg-16  (d) Vgg-19  (e) ResNet-50  (f) ResNet-101  (g) ResNeXt-101

**FIGURE 4.** Visualization of universal perturbations. The first row are universal perturbations generated by UAP and the second row are universal perturbations generated by Fast-UAP, the last row indicates the CNN models which are used to compute the corresponding universal perturbations. The pixel values are scaled for visibility.

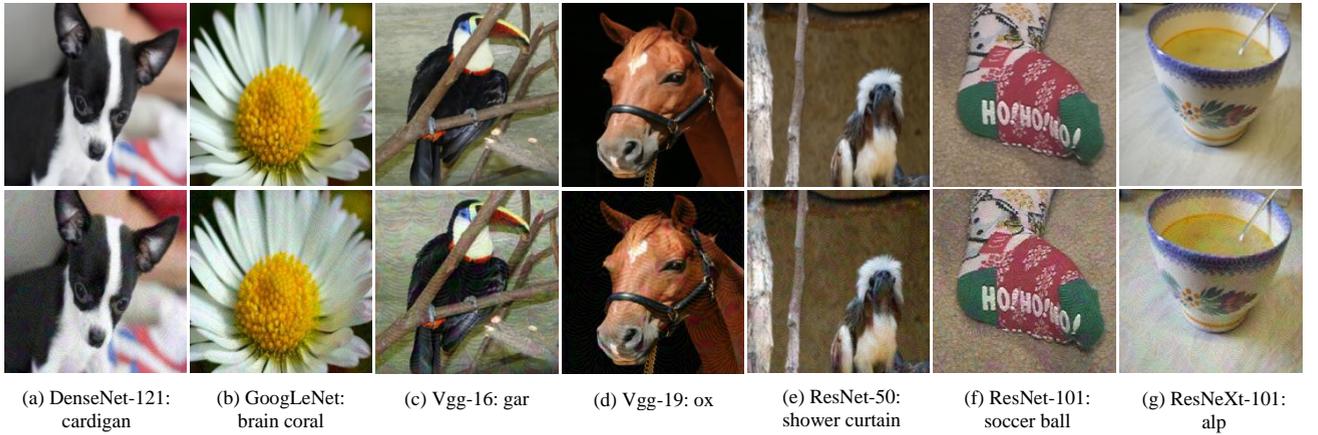

(a) DenseNet-121: cardigan  (b) GoogLeNet: brain coral  (c) Vgg-16: gar  (d) Vgg-19: ox  (e) ResNet-50: shower curtain  (f) ResNet-101: soccer ball  (g) ResNeXt-101: alp whether the current fooling rate reach the designated value $\delta$. The fooling rate $\mathcal{F}$ is computed as follows:

$$\mathcal{F} = \frac{1}{n}\sum_{i=1}^{n} r \text{ s.t. } r = \begin{cases} 1, & C(x_i + \bar{v}) \neq C(x_i) \\ 0, & C(x_i + \bar{v}) = C(x_i) \end{cases}.$$

**FIGURE 5.** Example of perturbed images. The first row are original images and the second row are perturbed images which are added universal perturbations generated by Fast-UAP, the last row indicates the CNN models which are used to compute the corresponding universal perturbations and the labels of perturbed images classified by GoogLeNet model. All original images are chosen from ILSVRC 2012 dataset.

Where $n$ is the number of testing samples. If $\mathcal{F} \geq \delta$, the algorithm terminates, or a new pass on training set $X$ will start to improve the quality of the universal perturbation. There will be a random shuffling to the set $X$ before every pass. The Fast-UAP algorithm is presented in detail in Algorithm 1.

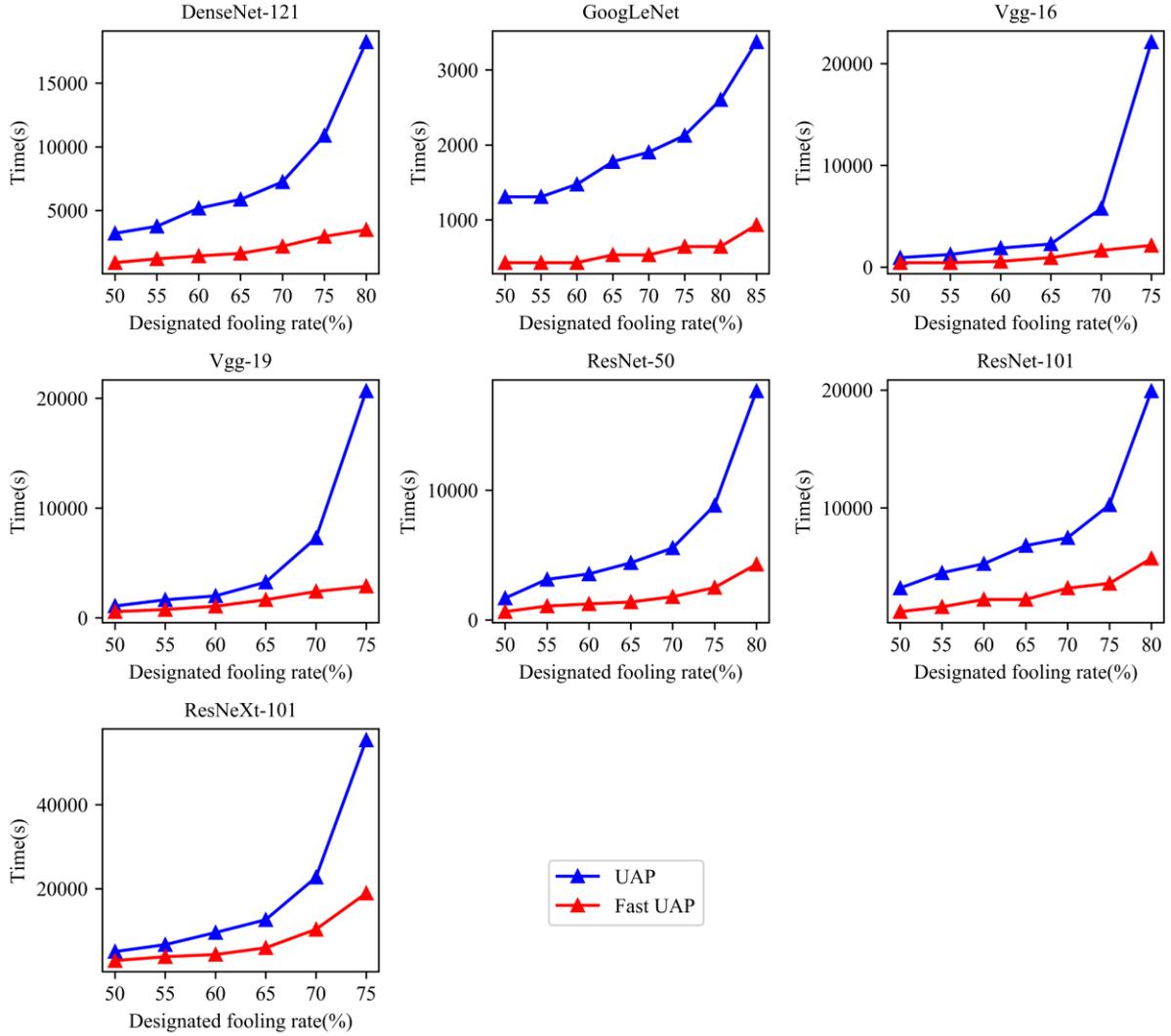

**FIGURE 6.** Time for generating universal perturbations to reach designated fooling rates by UAP and Fast-UAP.

The generated universal perturbations should give a $\ell_p$ norm restriction to be quasi-imperceptible. Though $\ell_\infty$ and $\ell_2$ norms are both common magnitude measurements of perturbations, we abandon $\ell_2$ norm because it does not limit every pixel values of perturbations and often causes noticeable chromatic aberration in perturbed images, as depicted in Fig. 3. A proper $\ell_\infty$ norm value should be chosen to make the perturbation imperceptible to human eyes, as well as make universal perturbations reaching a high fooling rate. In our algorithm, we set the maximum $\ell_\infty$ norm value of universal perturbations to be 10.

## IV. EXPERIMENTS

In this section, we describe two experiments to evaluate Fast-UAP. In the first experiment, we show that Fast-UAP generates universal perturbations faster than UAP to meet the same designated fooling rates. Then, we evaluate the fooling rates of universal perturbations generated by UAP and Fast-UAP under white-box attacks and black-box attacks respectively.

---

**Algorithm 1**: Fast-UAP algorithm to compute universal perturbations.

1: **input**: Preprocessed training set $X$, target classifier $C(\cdot)$ and designated fooling rate $\delta$.
2: **output**: universal perturbation vector $v$.
3: Initialize $v \leftarrow 0$.
4: **while** $\mathcal{F} < \delta$ **do**
5:     Shuffle training set $X$.
6:     **for** ordinal images $x_i$ in $X$ **do**
7:         **if** $C(x_i + v) = C(x_i)$ **then**
8:             **if** $v = 0$ **then**
9:                 Compute perturbation $\Delta_{v_i}$ that has the smallest magnitude and fools $C(\cdot)$:
10:                 $\Delta_{v_i} \leftarrow \arg\min_{v_i} \|v_i\|_2$ s.t. $C(x_i + v + v_i) \neq C(x_i)$.
11:             **else then**
12:                 Compute perturbation $\Delta_{v_i}$ that has similar orientation to $v$ and fools $C(\cdot)$:
13:                 $\Delta_{v_i} \leftarrow \arg\max_{v_i} \cos <v_i, v>$ s.t. $C(x_i + v + v_i) \neq C(x_i)$.
14:             **end if**
15:             Update the perturbation: $v \leftarrow v + \Delta_{v_i}$.
16:             Clip $v$ to maintain $\ell_\infty$ norm restriction.
17:         **end if**
18:     **end for**
19: **end while**

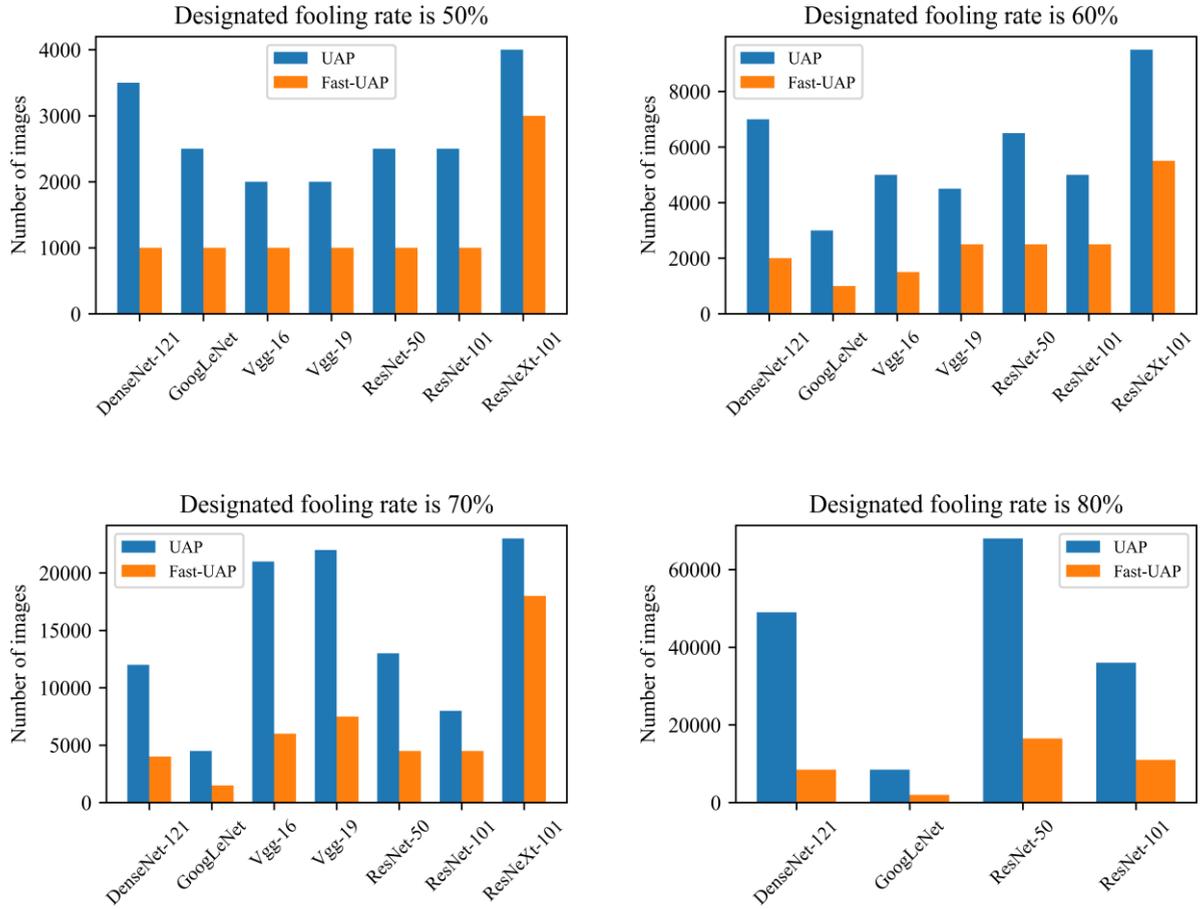

**FIGURE 7.** Number of images needed for generating universal perturbations to reach designated fooling rates by UAP and Fast-UAP. Universal perturbations generated by UAP for Vgg-16, Vgg-19, and ResNeXt-101 models cannot reach 80% fooling rate, so there is no comparison in the bottom right diagram for these models.

We pick 10000 random images as training samples from ILSVRC 2012 [28] training set, with 10 images for every class, and 50000 images from ILSVRC 2012 validation set as testing samples. To evaluate the performance of Fast-UAP and UAP on multiple CNN architectures, we choose seven popular CNN models such as DenseNet-121 [23], GoogLeNet [24], Vgg-16 [25], Vgg-19 [25], ResNet-50 [26], ResNet-101 [26], and ResNeXt-101 [27]. We generate universal perturbations for each of these CNN models via Algorithm 1. For ease of comparison, we also generate universal perturbations by UAP under the same experiment conditions. the size of input images and computed perturbations are 224×224 for ease of cross experiments. The visualization of universal perturbations generated by UAP and Fast-UAP for the aforementioned seven CNN models are shown in Fig. 4. The visual examples of perturbed images by adding universal perturbations generated by Fast-UAP are shown in Fig. 5. It can be seen that universal perturbations generated by Fast-UAP remain quasi-imperceptible to the human eye.

### A. THE PERFORMANCE OF FAST-UAP

To evaluate the performance of Fast-UAP, we now examine the time and images needed to generate universal perturbations to meet designated fooling rates by UAP and Fast-UAP respectively. We preset several fooling rates above 50%, then record time of UAP and Fast-UAP to generate universal perturbations for the seven aforementioned CNN models to meet one of the fooling rates and number of the corresponding images. We stop algorithm when fooling rates do not rise up because universal perturbations generated for different CNN models may have different fooling rate upper bounds. We run the experiments for each CNN model on the same equipment with no other jobs running on the system. The experiment results are shown in Fig. 6 and Fig. 7 respectively. It can be observed in Fig. 6 that Fast-UAP generates universal perturbations in a shorter time comparing with that of UAP on all tested CNN models. For example, for DenseNet-121 model, it takes UAP about 5100 seconds to generate a universal perturbation to meet the designated fooling rate of 60%, while it takes Fast-UAP about 1500 seconds to do so, which is an hour faster in time. It can also be seen from the diagrams in Fig. 7 that universal perturbations generated by Fast-UAP always reach a designated fooling rate with smaller number of training images comparing with that of UAP. For example, for ResNet-50 model, UAP needs about 6500 images to generate a universal perturbation to meet the designated fooling rate 60% while Fast-UAP only needs about 2500 images to do so.

TABLE I. Fooling rates of universal perturbations generated by UAP vs. by Fast-UAP. The percentages indicate fooling rates. The rows indicate the target models for which universal perturbations are computed, the fooling rate increments of Fast-UAP compared with that of UAP are also listed. The columns indicate test models. The diagonal entries in bold represents fooling rates under white-box attacks and other entries (except rightmost column) represent fooling rates under black-box attacks. The rightmost column indicates averages of each row.

|  |  | DenseNet-121 | GoogLeNet | Vgg-16 | Vgg-19 | ResNet-50 | ResNet-101 | ResNeXt-101 | Mean |
|---|---|---|---|---|---|---|---|---|---|
| Dense-Net-121 | UAP | **81.53%** | 58.25% | 46.18% | 43.79% | 52.37% | 52.06% | 47.56% | 54.53% |
|  | Fast-UAP | **90.67%** | 74.25% | 54.52% | 51.97% | 70.06% | 69.37% | 58.49% | 67.05% |
|  | Increment | **9.14%** | 16.00% | 8.34% | 8.18% | 17.69% | 17.31% | 10.93% | 12.51% |
| Goog-LeNet | UAP | 21.63% | **87.04%** | 20.73% | 20.88% | 23.12% | 22.32% | 17.37% | 30.44% |
|  | Fast-UAP | 29.58% | **94.88%** | 27.67% | 27.51% | 31.95% | 29.98% | 22.18% | 37.68% |
|  | Increment | 7.95% | **7.84%** | 6.94% | 6.63% | 8.83% | 7.66% | 4.81% | 7.24% |
| Vgg-16 | UAP | 57.82% | 65.51% | **77.31%** | 69.54% | 55.54% | 57.03% | 47.26% | 61.43% |
|  | Fast-UAP | 65.19% | 76.78% | **84.46%** | 77.67% | 63.82% | 66.04% | 52.77% | 69.53% |
|  | Increment | 7.37% | 11.27% | **7.15%** | 8.13% | 8.28% | 9.01% | 5.51% | 8.10% |
| Vgg-19 | UAP | 54.56% | 63.94% | 69.99% | **76.12%** | 53.77% | 52.30% | 41.44% | 58.87% |
|  | Fast-UAP | 59.52% | 72.20% | 76.24% | **80.40%** | 61.41% | 60.91% | 48.01% | 65.53% |
|  | Increment | 4.96% | 8.26% | 6.25% | **4.28%** | 7.64% | 8.61% | 6.57% | 6.65% |
| ResNet-50 | UAP | 53.7% | 59.41% | 43.44% | 42.97% | **81.41%** | 63.15% | 46.81% | 55.84% |
|  | Fast-UAP | 65.93% | 68.24% | 48.82% | 47.21% | **89.58%** | 72.93% | 52.90% | 63.66% |
|  | Increment | 12.23% | 8.83% | 5.38% | 4.24% | **8.17%** | 9.78% | 6.09% | 7.82% |
| ResNet-101 | UAP | 50.96% | 56.49% | 44.63% | 43.48% | 59.56% | **83.83%** | 41.73% | 54.38% |
|  | Fast-UAP | 62.35% | 66.44% | 49.52% | 49.38% | 71.28% | **90.67%** | 52.61% | 63.18% |
|  | Increment | 11.39% | 9.95% | 4.89% | 5.90% | 11.72% | **6.84%** | 10.88% | 8.80% |
| Res-NeXt-101 | UAP | 56.45% | 44.71% | 34.93% | 34.39% | 47.12% | 50.93% | **77.85%** | 49.48% |
|  | Fast-UAP | 64.45% | 63.69% | 48.91% | 46.66% | 62.38% | 63.99% | **80.91%** | 61.57% |
|  | Increment | 8.00% | 18.98% | 13.98% | 12.27% | 15.26% | 13.06% | **3.06%** | 12.09% |

### B. THE FOOLING RATE UNDER WHITE-BOX ATTACKS AND BLACK-BOX ATTACKS SCENARIOS

In the last experiment, we empirically observed that universal perturbations generated by Fast-UAP always achieve higher fooling rates. To evaluate fooling performance of Fast-UAP more completely, we now examine the fooling rates of UAP and Fast-UAP under white-box attacks, where attacker can access to victim model, as well as under black-box attacks, where attacker have little knowledge about victim model. That is, for one instance of universal perturbation generated for a target model (e.g. DenseNet-121), we study its fooling rates on the same target model as well as other models (e.g. GoogLeNet) which are not used to generate it. We set designated fooling rate δ to be 100% to get the actual maximum fooling rates of universal perturbations and terminate algorithm when fooling rates does not rise up. The experiment results on testing samples (they are not used to generate the universal perturbations) for the aforementioned seven CNN models are shown in Table. I. All diagonal entries in bold in the table are fooling rates of white-box attacks while other entries except the right most column represent fooling rates of black-box attacks. It is empirically observed from the table that universal perturbations generated by Fast-UAP achieve an average fooling rate increment of 6.64% under white-box attacks. In particular, for DenseNet-121 model, the fooling rate increment is even higher than 9%. As for black-box attacks, we can also find that fooling rates of universal perturbations generated by Fast-UAP is always higher than that of UAP. This experiment shows that Fast-UAP has better fooling performance in both white-box attacks and black-box attacks.

### V. CONCLUTION

Though deep neural networks show state-of-the-art performance in diverse tasks, the vulnerability of deep neural networks has been the hot topic in recent years. Specially, there is an iterative algorithm called UAP to generate one single perturbation to make a set of natural images to be misclassified on CNN models with high probability. In this paper, we illustrated the orientations of perturbation vectors have influence on magnitude growth of aggregation of these vectors, which may slow down the process to generate universal perturbations. Then we proposed Fast-UAP to overcome this influence. The experiments showed that Fast-UAP can generate universal perturbations much faster. Furthermore, we empirically observed in experiments that the universal perturbations generated by Fast-UAP achieve higher fooling rates in both white-box attacks and black-box attacks. A research to explore why aggregating perturbations with a similar orientation leads to an increment in the fooling rate will be the focus of our future work.